\documentclass[lettersize,journal]{IEEEtran}
\usepackage{amssymb}
\usepackage{float}
\usepackage{hyperref}
\usepackage{booktabs,multirow}
\usepackage{longtable}%
\usepackage{booktabs}%
\usepackage{tabularx}
\usepackage{makecell}
\usepackage{amsmath,amsfonts}
\usepackage{algorithmic}
\usepackage{algorithm}
\usepackage{array}
\usepackage[caption=false,font=normalsize,labelfont=sf,textfont=sf]{subfig}
\usepackage{textcomp}
\usepackage{stfloats}
\usepackage{url}
\usepackage{verbatim}
\usepackage{graphicx}
\usepackage{cite}
\usepackage{orcidlink}
\usepackage[english]{babel}
\usepackage{csquotes}
\begin{document}
\bstctlcite{IEEEexample:BSTcontrol}

\title{A Review of Multimodal Explainable Artificial Intelligence: Past, Present and Future}

\author{Shilin Sun\textsuperscript{\orcidlink{0009-0007-8897-3737}},Wenbin An\textsuperscript{\orcidlink{0000-0003-0062-7201}},Feng Tian\textsuperscript{\orcidlink{0000-0001-7888-0587}}~\IEEEmembership{Senior Member, IEEE}, Fang Nan\textsuperscript{\orcidlink{0000-0002-5896-0237}}, Qidong Liu\textsuperscript{\orcidlink{0000-0002-0751-2602}}, 
Jun Liu\textsuperscript{\orcidlink{0000-0002-6004-0675}}~\IEEEmembership{Senior Member, IEEE}, Nazaraf Shah\textsuperscript{\orcidlink{0000-0002-4140-8200}} and Ping Chen\textsuperscript{\orcidlink{0000-0003-3789-7686}}
\thanks{This work was supported by National Science and Technology Major Project (2022ZD0117102), National Natural Science Foundation of China (62293551, 62177038, 62277042, 62137002, 61937001,62377038). Project of China Knowledge Centre for Engineering Science and Technology, ``LENOVO-XJTU’’ Intelligent Industry Joint Laboratory Project. We sincerely thank Chenyang Wang, Yuqi Sun, Xinyue Shen, Zhenzhen Xuan, Kexuan Li, Jiayuan Li, Haonan Miao, Zengyi Chen, Yan Li and Zhihao Jiang for their help with literature organization and data collection. We also appreciate the assistance of Di Zhang and Zhi Zeng in reviewing the content. (Corresponding author: Feng Tian).} 

\thanks{Shilin Sun and Feng Tian  are with the School of Computer Science and Technology, Xi’an Jiaotong University, Xi’an, 710049, China, and also with the  Ministry of Education Key Laboratory of Intelligent Networks and Network Security, Xi’an Jiaotong University, Xi’an, 710049, China (e-mail: shilinsun@stu.xjtu.edu.cn; fengtian@mail.xjtu.edu.cn).}

\thanks{Wenbin An, Fang Nan, and Qidong Liu are with the Faculty of Electronic and Information Engineering, Xi’an Jiaotong University, Xi’an, 710049, China, and also with the Shaanxi Province Key Laboratory of Big Data Knowledge Engineering, Xi’an Jiaotong University, Xi’an, 710049, China  (e-mail: wenbinan@stu.xjtu.edu.cn; nanfangalan@gmail.com; liuqidong@stu.xjtu.edu.cn).}

\thanks{Jun Liu is with the School of Computer Science and Technology, Xi’an Jiaotong University, Xi’an, 710049, China, and also with the  Shaanxi Province Key Laboratory of Big Data Knowledge Engineering, Xi’an Jiaotong University, Xi’an, 710049, China (email: liukeen@xjtu.edu.cn)}

\thanks{Nazaraf Shah is with the Institute for Future Transport and Cities, Coventry University, Priory Street, Coventry, CV1 5FB, United Kingdom (e-mail: aa0699@coventry.ac.uk ).}

\thanks{Ping Chen is with the Department of Engineering, University of Massachusetts Boston, Boston, MA 02125 USA (e-mail: Ping.Chen@umb.edu)}}

\markboth{Journal of \LaTeX\ Class Files,~Vol.~14, No.~8, August~2021}%
{Shell \MakeLowercase{\textit{et al.}}: A Sample Article Using IEEEtran.cls for IEEE Journals}


\maketitle

\begin{abstract}
Artificial intelligence (AI) has rapidly developed through advancements in computational power and the growth of massive datasets. However, this progress has also heightened challenges in interpreting the ``black-box'' nature of AI models. To address these concerns, eXplainable AI (XAI) has emerged with a focus on transparency and interpretability to enhance human understanding and trust in AI decision-making processes. In the context of multimodal data fusion and complex reasoning scenarios, the proposal of Multimodal eXplainable AI (MXAI) integrates multiple modalities for prediction and explanation tasks. Meanwhile, the advent of Large Language Models (LLMs) has led to remarkable breakthroughs in natural language processing, yet their complexity has further exacerbated the issue of MXAI. To gain key insights into the development of MXAI methods and provide crucial guidance for building more transparent, fair, and trustworthy AI systems, we review the MXAI methods from a historical perspective and categorize them across four eras: traditional machine learning, deep learning, discriminative foundation models, and generative LLMs. We also review evaluation metrics and datasets used in MXAI research, concluding with a discussion of future challenges and directions. A project related to this review has been created at \url{https://github.com/ShilinSun/mxai_review}.
\end{abstract}.

\begin{IEEEkeywords}
 large language models (LLMs), multimodal explainable
artificial intelligence (MXAI), historical perspective, generative.
\end{IEEEkeywords}

\section{Introduction}
\begin{figure}[!t]
\centering
\includegraphics[width=3.5in]{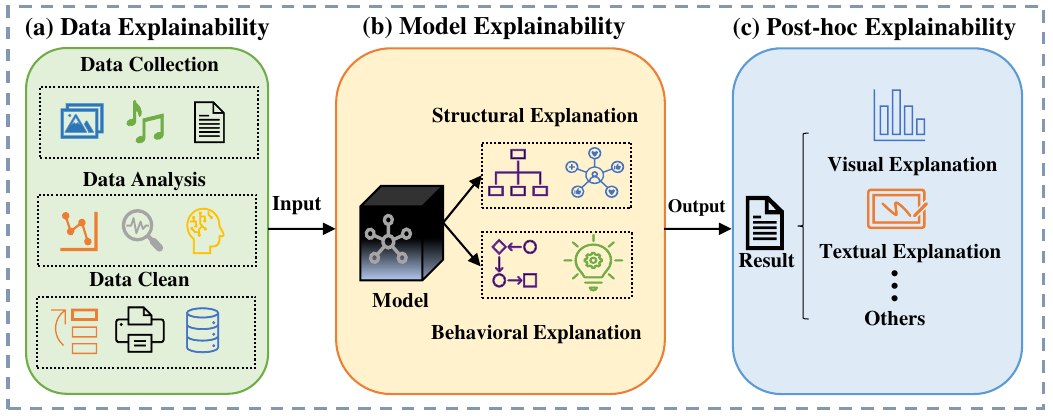}
\caption{Illustrative diagram of Multimodal Explainable Artificial Intelligence.}
\label{fig1}
\end{figure}

\IEEEPARstart{A}{dvancements} in AI have significantly impacted computer science, with works like Transformer~\cite{vaswani2017attention}, BLIP-2~\cite{li2023blip} and  ChatGPT~\cite{openai2022chatgpt} excelling in natural language processing (NLP), computer vision, and multimodal tasks by integrating diverse data types. The development of related technologies has driven the advancement of specific applications. For example, in autonomous driving, systems need to integrate data from various sensors, including vision, radar, and LiDAR, to ensure safe operation in complex road environments~\cite{Sachdeva_2024_WACV}. Similarly, health assistants require transparency and trustworthiness to be easily understood and verified by both doctors and patients~\cite{yang2022unbox}. Understanding how these models combine and interpret different modalities is crucial for enhancing model credibility and user trust. Moreover, increasing model scales pose challenges in computational cost, interpretability, and fairness, driving the demand for Explainable AI (XAI)~\cite{rodis2023multimodal}. As models, including generative LLMs, become increasingly complex and data modalities more diverse, single-modal XAI methods can no longer meet user demands. Therefore, Multimodal eXplainable AI (MXAI) addresses these challenges by utilizing multimodal data in either the model’s prediction or explanation tasks, as shown in Fig.~\ref{fig1}. We categorize MXAI  into three types based on the data processing sequence: data explainability (pre-model), model explainability (in-model), and post-hoc explainability (post-model). In multimodal prediction tasks, the model processes multiple data modalities, such as text, images, and audio. In multimodal explanation tasks, various modalities are used to explain the results, offering a more comprehensive explanation of the final output.

\begin{figure}[!t]
\centering
\includegraphics[width=3.5in]{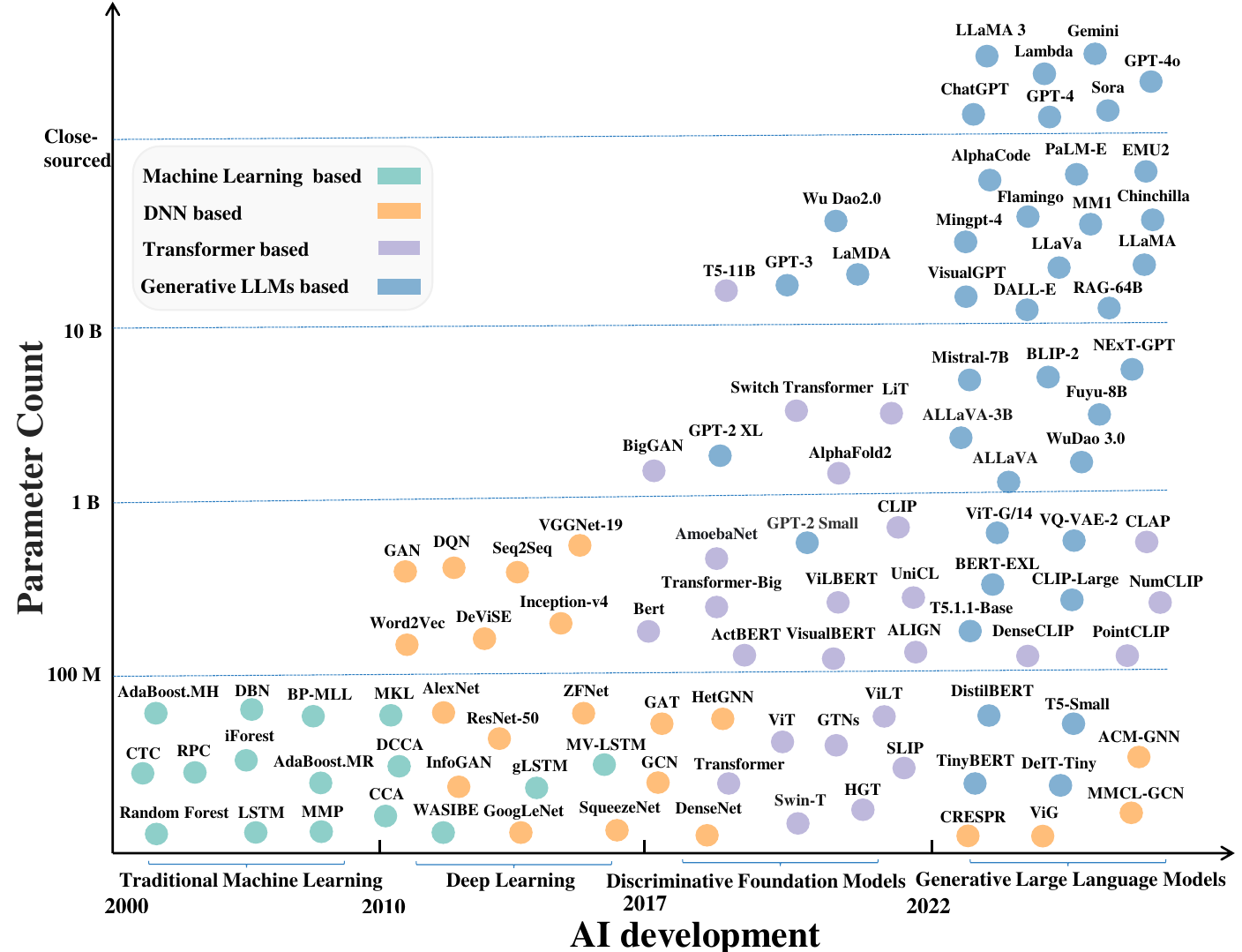}
\caption{As AI advances, increasing computational power has led to larger model parameter counts and a growing number of multimodal models.}
\label{fig2}
\end{figure}

To review the history of MXAI and anticipate its development, we first categorized different periods and retrospectively examined various models from a historical perspective (as illustrated in Fig.~\ref{fig2}). During the traditional machine learning era (2000-2009), the availability of limited structured data favored interpretable models like decision trees. In the deep learning era (2010-2016), the advent of large annotated datasets, such as ImageNet~\cite{deng2009imagenet}, coupled with increased computational power, led to the rise of complex models and explainable studies, including visualizing neural network kernels~\cite{erhan2009visualizing}. In the discriminative foundation models era (2017-2021), the emergence of Transformer models, leveraging large-scale text data and self-supervised learning, revolutionized NLP. This shift sparked significant research into interpreting attention mechanisms~\cite{vaswani2017attention,chen2022beyond,chefer2021generic,chefer2021transformer}. In the generative large language models era (2022-2024), the integration of vast multimodal data is driving the development of generative Large Language Models (LLMs), such as ChatGPT~\cite{openai2022chatgpt}, and multimodal fusion techniques. These advancements provide comprehensive explanations, enhancing model transparency and trust. This evolution has led to a focus on MXAI, which interprets models handling diverse data types~\cite{rodis2023multimodal}.
\IEEEpubidadjcol

However, recent reviews on XAI often overlook historical developments and focus mainly on unimodal approaches. For example, while~\cite{rodis2023multimodal} categorizes MXAI methods by modality count, explanation stages, and method types, it misses explainability techniques for LLMs. Although Ali et al.~\cite{RN7} propose a comprehensive four-axis taxonomy, it lacks multimodal and LLMs related summaries. However, reviews like~\cite{singh2024rethinking, RN152}, and~\cite{luo2024understanding} focus solely on explainability for LLMs.
Our study addresses these shortages by offering a \textbf{historical perspective} of \textbf{MXAI}. And we categorize MXAI methods into four eras (traditional machine learning, deep learning, discriminative foundation models, and generative large language models), and each is divided into three categories (data, model, and post-hoc explainability). The key novel contributions of this paper are concluded as follows:
\begin{itemize}
    \item We offer a historical summary and analysis of MXAI methods, including both traditional machine learning and current MXAI methods based on LLMs.
    \item We analyze methods across eras, covering data, model, and post-hoc explainability, along with relevant datasets, evaluation metrics, future challenges and directions.
    \item We review existing methods, summarize current approaches, and offer insights into future developments and a systematic and comprehensive perspective from the standpoint of historical evolution.
\end{itemize}
\section{Preliminaries}

\subsection{MXAI Method Taxonomy}
We categorize Artificial Intelligence (AI) into four eras based on key technological milestones: the release of the ImageNet dataset in 2009~\cite{deng2009imagenet}, marking deep learning's rise over traditional machine learning, the introduction of the Transformer model in 2017~\cite{vaswani2017attention}, distinguishing deep learning from discriminative models, and the advent of ChatGPT in 2022~\cite{openai2022chatgpt}, ushering in the era of generative large models.
\begin{itemize}
    \item \textbf{The Era of Traditional Machine Learning (2000-2009)}: In this era, MXAI methods are characterized by manual feature engineering and rule-based systems, which ensure transparent interpretability~\cite{peng2005feature,nunez2006rule}. Decision trees and basic visualization tools further support interpretability, with evaluations, focused on accuracy and human-centered metrics~\cite{nefesliouglu2010assessment,domingos2000mining}. Despite their advantages, these methods are limited by scalability and flexibility issues due to small datasets and modality constraints, thus setting the stage for the development of more advanced MXAI methods in subsequent years.
    \item \textbf{The Era of Deep Learning (2010-2016)}: MXAI methods leverage deep neural networks to process and interpret multimodal data. Key characteristics include integrating different data types, developing multimodal architectures that combine Convolutional Neural Networks (CNNs) and Recurrent Neural Networks (RNNs), and focusing on joint representation learning~\cite{kingma2013auto,higgins2017beta}. This era lays the groundwork for modern MXAI by advancing data integration and interpretability techniques.
    \item \textbf{The Era of Discriminative Foundation Models (2017-2021)}: This era is characterized by the development and widespread adoption of models like Transformer~\cite{vaswani2017attention}, CLIP~\cite{radford2021learning}, and their variants. During this era, MXAI methods focus on enhancing the interpretability of these models by leveraging multimodal data, such as text, images, and audio. Researchers aim to explain model decisions through techniques like attention visualization, feature attribution, and gradient-based methods. This era features advancements in multimodal inputs with foundation models, providing comprehensive explanations and improving the trustworthiness of models.
    \item \textbf{The Era of Generative Large Language Models (2022-2024)}: In this era, MXAI methods evolve to address the distinct challenges and opportunities presented by generative LLMs like ChatGPT~\cite{openai2022chatgpt}. Unlike the more transparent Transformer-based discriminative models, LLMs often restrict direct access, necessitating innovative interpretability methods. These techniques leverage the interactivity of LLMs to enhance adaptive explanations, manage the complexity of multimodal data and outputs, and address ethical concerns such as biases. The focus is on developing sophisticated, ethical, and user-responsive MXAI methods to elucidate the intricate reasoning and decision-making processes of these models.
\end{itemize}

In each era, we categorize MXAI methods into three main categories: \textbf{data}, \textbf{model}, and \textbf{post-hoc} explainability. As shown in Fig.~\ref{fig1}, data explainability provides insights by summarizing and analyzing data, facilitating feature engineering and standardization, such as data collection, data analysis and data clean. For example, Principal Component Analysis (PCA)~\cite{jolliffe2002principal} can be used to reduce dimensionality and orthogonalize features across multiple modalities, such as text, images, and audio. Model explainability reveals the internal structure and algorithms by structural explanation, behavioral explanation and so on. For example, CLIP~\cite{radford2021learning} can provide explainability through attention maps, which show how different modalities (e.g., text and images) contribute to the model's decision-making process by highlighting key features from each modality. Post-hoc explainability refers to methods used to explain the model's predictions after the model has been trained. For example, Grad-CAM++~\cite{chattopadhay2018grad} refines Grad-CAM by offering more precise visual explanations for models like trained CLIP. It enhances the localization of image regions associated with specific text for clearer insights into multimodal alignment.
\subsection{Comparison with Previous Surveys}
\begin{table*}[!t]
\caption{Four-year XAI survey summary and comparison} 
\label{tab1}
\centering
\resizebox{\textwidth}{!}{\begin{tabular}{c|c|c|c|c|c|c|c|c|c|c|c|c}
\hline
 \rotatebox{90}{\textbf{Ref.}}&\rotatebox{90}{\textbf{Published Year}} & \rotatebox{90}{\textbf{Literature coverage range}} &\rotatebox{90}{\textbf{Existing surveys are analyzed}} &\rotatebox{90}{\textbf{MXAI methodology}} & \rotatebox{90}{\textbf{Transformer explainability}} & \rotatebox{90}{\textbf{LLMs explainability}}&\rotatebox{90}{\textbf{Historical perspective}}&\rotatebox{90}{\textbf{Data explainability}}&\rotatebox{90}{\textbf{Model explainability}}&\rotatebox{90}{\textbf{Post-hoc explainability}}& \rotatebox{90}{\textbf{Evaluation methods}} &\textbf{Main theme}\\
 \hline
Ours& 2024 & 2000-2024&\checkmark&\checkmark &\checkmark& \checkmark&\checkmark &\checkmark& \checkmark&\checkmark &\checkmark&Historical perspective MXAI\\
\cline{1-13}
~\cite{RN13}& 2023 & 2016-2023&&\checkmark
&&&&&$\checkmark$&$\checkmark$ &$\checkmark$&MXAI methodology\\
\cline{1-13}
\cline{1-13}
\cline{1-13}
\hline
~\cite{RN157}& 2024 &2016-2024&&
&\checkmark&\checkmark&&&&\checkmark&&Explainable Generative AI\\
\cline{1-13}
~\cite{RN152}& 2024 &2017-2024&&
&\checkmark&\checkmark&&&&\checkmark&\checkmark&LLM explainability strategies\\
\cline{1-13}
~\cite{RN149}& 2024 &2017-2024&&
&\checkmark&\checkmark&&&&\checkmark&\checkmark&LLMs explainability\\
\cline{1-13}
~\cite{singh2024rethinking}& 2024 &2017-2024&&
&&\checkmark&&\checkmark&&\checkmark&\checkmark&LLMs explainability\\
\cline{1-13}
~\cite{luo2024understanding}& 2024 &2016-2023&&
&&\checkmark&&&&\checkmark&\checkmark&LLMs explainability\\
\cline{1-13}
~\cite{RN12}& 2023 & 2008-2022&\checkmark&
&&&&&&$\checkmark$ &\checkmark&User and their concerns\\
\cline{1-13}
~\cite{RN10}& 2023 & 2018-2022&&
&&&&&$\checkmark$&$\checkmark$ &&XAI applications\\
\cline{1-13}
~\cite{kashefi2023explainability}& 2023 & 2016-2023&&
&$\checkmark$&&&&$\checkmark$&$\checkmark$ &$\checkmark$&Vision Transformer explainability\\
\cline{1-13}
~\cite{RN32}& 2023 & 2016-2023&&
&$\checkmark$&&&&$\checkmark$&$\checkmark$ &$\checkmark$&Vision Transformer explainability\\
\cline{1-13}
~\cite{RN181}& 2023 & 2018-2022&$\checkmark$&
&&&&&$\checkmark$&$\checkmark$ &$\checkmark$&Human-centered XAI\\
\cline{1-13}
~\cite{RN8}& 2023 & 2017-2021&$\checkmark$&
&&&&&$\checkmark$&$\checkmark$ &$\checkmark$&Challenges and trends in XAI\\
\cline{1-13}
~\cite{RN7}& 2022 & 2016-2022&\checkmark&
&&&& \checkmark&\checkmark&\checkmark &\checkmark&Model’s trustworthiness\\
\cline{1-13}
~\cite{cambria2023survey}& 2022 & 2006-2021&&
&&&&& \checkmark&\checkmark &&Natural Language Explanations\\
\cline{1-13}
~\cite{tiddi2022knowledge}& 2022 & 2015-2020&&
&&&&&& &&Knowledge based XAI\\
\cline{1-13}
~\cite{holzinger2022explainable}& 2022 & 2016-2021&&
&&&&&\checkmark&\checkmark &&Introduction to XAI\\
\cline{1-13}
~\cite{guidotti2022counterfactual}& 2022 & 2017-2022&$\checkmark$&
&&&&&&\checkmark &\checkmark&Counterfactual explanations
\\
\cline{1-13}
~\cite{theissler2022explainable}& 2022 & 2018-2022&$\checkmark$&
& &&&&&\checkmark &\checkmark&XAI for time series
\\
\cline{1-13}
~\cite{yang2022unbox}& 2022 & 2018-2021&$\checkmark$&
&&&&&&\checkmark &&XAI in healthcare
\\
\cline{1-13}
~\cite{stepin2021survey}& 2021 & 1991-2020&\checkmark&
&&&&&\checkmark&\checkmark &\checkmark&Contrastive and Counterfactual XAI

\\
\cline{1-13}
~\cite{vilone2021notions}& 2021 & 2015-2020&\checkmark&
&&&&&& &\checkmark&Evaluation approaches of XAI
\\
\cline{1-13}
~\cite{meske2022explainable}& 2021 & 2017-2020&&
&&&&&& &&Black-box issue
\\
\cline{1-13}
~\cite{linardatos2020explainable}& 2021 & 2016-2020&&
&&&&&\checkmark& &&ML interpretability methods
\\
\cline{1-13}
~\cite{islam2021explainable}& 2021 & 2016-2020&\checkmark&
&&&&&&\checkmark &&XAI methods classification
\\
\cline{1-13}
~\cite{vassiliades2021argumentation}& 2021 & 2014-2020&\checkmark&
&&&&&& &&Argumentation enabling XAI
\\
\cline{1-13}
~\cite{burkart2021survey}& 2021 & 2015-2020&&
&&&&\checkmark&&\checkmark&\checkmark&XAI methods classification
\\
\cline{1-13}
~\cite{Gerlings2021}& 2021 & 2016-2020&&
&&&&&& &&Necessity of explainability
\\
\cline{1-13}
~\cite{hussain2021explainable}& 2021 & 2017-2020&&
&&&&&& &&User and their concerns
\\
\cline{1-13}
~\cite{li2022interpretable}& 2021 & 2016-2020&&
&&&&&\checkmark&\checkmark&\checkmark&XAI methods classification
\\
\hline
\end{tabular}}
\end{table*}
Based on~\cite{RN7}, we present the latest and most comprehensive comparison of XAI-related reviews from the past four years XAI-related reviews in the last four years as Table~\ref{tab1}. Despite numerous XAI reviews published over the last five years (see Table~\ref{tab1}), some are mostly limited to specific time eras or focused on XAI applications for particular technologies.
For example, recent XAI researchs~\cite{kashefi2023explainability,RN32} have largely concentrated on summarizing developments after Transformer and has predominantly focused on unimodal explainability. There is a notable gap in exploring the interpretability of LLMs from a historical and developmental perspective, as well as leveraging multimodal information. While Rodis et al.~\cite{RN13} review MXAI methods in terms of both models and post-hoc interpretability, it primarily addresses work post-2016 excluding LLMs and does not cover the explainability of Transformer and their variants.

In summary, as shown in Table~\ref{tab1}, XAI methods have evolved with advancements in artificial intelligence, with Transformer and large model interpretability emerging from this progression. As multimodal data becomes more prominent, MXAI methods are gaining increasing attention~\cite{RN13}. This paper addresses shortages in existing reviews by offering a comprehensive historical overview of MXAI methods, segmented into four eras: traditional machine learning (See \hyperlink{se3}{Section III}), deep learning (See \hyperlink{se4}{Section IV}), discriminative foundation models (See \hyperlink{se5}{Section V}), and generative large language models (See \hyperlink{se6}{Section VI}). Each era is examined with respect to data, models, and post-hoc explainability, and we also summarize relevant datasets and evaluation metrics for assessing MXAI methods (See \hyperlink{eval}{Section VII}). In \hyperlink{se8}{Section VIII}, we discuss the future challenges and directions in MXAI and conclude with a summary.
\hypertarget{se3}{}
\section{The era of Traditional Machine Learning}
As shown in Table~\ref{tab2}, data explainability in this era focuses primarily on data dimensionality reduction, including feature selection and extraction methods. Model explainability mainly involves shallow machine learning models, such as decision trees and Bayesian models. Post-hoc explainability techniques mainly are model-agnostic and model-specific approaches. These methods lay a crucial foundation for multimodal explainability research, with their techniques and insights extended to multimodal contexts.
\begin{table*}[!t]
\caption{Summary of MXAI methods in the traditional machine learning era}
\label{tab2}
\centering
\begin{tabular}{c|c|c|cc}
\hline
\multirow{14}{*}{\makecell[c]{The Era of Traditional\\Machine Learning (2000-2010)}} & \multirow{4}{*}{\hyperlink{Data explainability}{Data explainability}} & \multicolumn{1}{c|}{\multirow{3}{*}{Feature selection methods}} & \multicolumn{1}{c|}{Filter} & {\cite{peng2005feature,ding2005minimum,biesiada2008feature,horlings2008emotion,estevez2009normalized,huang2010optimization,bulling2010eye} } \\ \cline{4-5} 
 &  & \multicolumn{1}{c|}{} & \multicolumn{1}{c|}{Wrapper} &{\cite{cotter2001backward,guyon2002gene,colak2003feature,bensch2005feature}}  \\ \cline{4-5} 
 &  & \multicolumn{1}{c|}{} & \multicolumn{1}{c|}{Embedded} & \cite{breiman2001random,osei2003exploration,yan2004ant,shazzad2005optimization,huang2006wrapper,tan2006hybrid,fatourechi2007application,gheyas2010feature} \\ \cline{3-5} 
 &  & \multicolumn{1}{r|}{Feature extraction methods} & \multicolumn{2}{c}{\cite{hyvarinen2000independent,tenenbaum2000global,roweis2000nonlinear,martinez2001pca,jolliffe2002principal,ye2004two,sanguansat2006two,van2008visualizing,liang2008note,zhi2008two,janecek2008relationship, ringner2008principal,zheng20081d,ji2008generalized,lu2008mpca,wright2009robust,jenatton2010structured,zhang2010two,wang2010unsupervised}} \\ \cline{2-5} 
 & \multirow{5}{*}{\hyperlink{Model explainability}{Model explainability}} & Linear/logistic regression & \multicolumn{2}{c}{\cite{jaccard2001interaction,bursac2008purposeful,peng2002introduction,mood2010logistic}} \\ \cline{3-5} 
 &  & Decision Tree & \multicolumn{2}{c}{\cite{nefesliouglu2010assessment,domingos2000mining,sing2005rocr,ramos2005induction,ryu2008supporting,rajesh2009learning}} \\ \cline{3-5} 
 &  & K-Nearest Neighbors & \multicolumn{2}{c}{\cite{guo2004k,li2004application}} \\ \cline{3-5} 
 &  & Rule-based learning & \multicolumn{2}{c}{\cite{nunez2006rule,johansson2004truth,nunez2002rule}} \\ \cline{3-5} 
 &  & Bayesian models & \multicolumn{2}{c}{\cite{l2008bayesian,neelon2010bayesian,min2007probabilistic,koop2007bayesian}} \\ \cline{2-5} 
 & \multirow{5}{*}{\hyperlink{Causal Inference}{Post-hoc explainability}} & \multirow{3}{*}{Model-agnostic techniques} & \multicolumn{1}{c|}{Causal Explanation} & \cite{Gefen,woodward2005making,keil2006explanation,pearl2009causal} \\ \cline{4-5} 
 &  &  & \multicolumn{1}{c|}{Causal Prediction} &\cite{berk2008statistical,shmueli2011predictive,van2007interplay,RN101}  \\ \cline{4-5} 
 &  &  & \multicolumn{1}{c|}{Causal Intervention} &\cite{dawid2000causal,victora2004evidence,korb2004varieties,hagmayer2007causal,hudgens2008toward,pearl2009causality}  \\ \cline{3-5} 
 &  & \multirow{2}{*}{Model-specific techniques} & \multicolumn{1}{c|}{Tree ensemble} &\cite{domingos2000mining,sing2005rocr,ramos2005induction,ryu2008supporting,rajesh2009learning,chebrolu2005feature,chen2006building,eruhimov2008transferring,ram2011density,ankerst2000towards,lavravc2007data,sandri2008bias,menze2009comparison,nguyen2002visualization,breiman2001random,sandri2010analysis,ankerst1999visual}  \\ \cline{4-5} 
 &  &  & \multicolumn{1}{c|}{Support vector machines} & \cite{barakat2008eclectic,fu2004extracting,nunez2002support,zhang2005rule,fung2005rule,chen2007multiple,navia2006support} \\ \hline
\end{tabular}
\end{table*}
\hypertarget{Data explainability}{}
\subsection{Data explainability}
With advancements in technology, the proliferation of data generation has led to the emergence of high dimensionality issues, commonly referred to as dimensional catastrophe. To address this, explainability methods based on dimensionality reduction aim to reduce redundancy and computational load. Among these methods, feature selection retains original features and interpretability by selecting relevant subsets. In contrast, feature extraction creates new features through mathematical transformations, enhancing processing efficiency but potentially reducing interpretability.

Feature selection methods are used to choose important features from a dataset and are categorized into filter, wrapper, and embedded methods. Filter methods, like those based on mutual information, assess each feature's relevance independently of learning algorithms~\cite{peng2005feature}. Wrapper methods, exemplified by Recursive Feature Elimination (RFE)~\cite{guyon2002gene}, evaluate subsets based on model performance degradation. Embedded methods, integrated into training processes like tree-based models, determine feature importance, improving transparency in model decision-making~\cite{breiman2001random}.

Feature extraction methods transform original data into a low-dimensional feature space, compressing data complexity. Unlike feature selection methods, it creates new features rather than selecting existing ones. Principal Component Analysis (PCA)~\cite{jolliffe2002principal} determines principal components by eigenvectors of the data's covariance matrix, enabling dimensionality reduction and data structure visualization. Linear Discriminant Analysis (LDA)~\cite{ye2004two} maximizes inter-class distance and minimizes intra-class variance, enhancing data interpretability. t-SNE~\cite{van2008visualizing} preserves local structure between data points in high-dimensional space to handle modal differences. UKDR~\cite{wang2010unsupervised} extends kernel dimensionality reduction for unsupervised learning, offering superior low-dimensional embeddings compared to t-SNE and PCA.
\hypertarget{Model explainability}{}
\subsection{Model explainability}
Logistic Regression (LR) is used for binary classification, while Linear Regression is its counterpart for continuous outcomes. Both models assume linear relationships between predictors and outcomes, limiting flexibility but ensuring transparency. Most authors agree on the robustness of using different techniques to analyze and express LR~\cite{jaccard2001interaction,bursac2008purposeful,peng2002introduction}.
Techniques from other disciplines, like visualization, effectively interpret and present regression model results to non-statistical users~\cite{mood2010logistic}.
Apart from this transparent models are Decision Trees~\cite{nefesliouglu2010assessment}, K-Nearest Neighbors~\cite{guo2004k,li2004application}, Rule-based learning~\cite{johansson2004truth,nunez2002rule}, Bayesian models~\cite{l2008bayesian,neelon2010bayesian,min2007probabilistic,koop2007bayesian}, etc.
\hypertarget{Causal Inference}{}\subsection{Post-hoc explainability}
Model-agnostic approaches, such as causal inference methods, have emerged as leading post-hoc explanation techniques in this era due to their model independence, focus on causal relationships, data-driven nature, improvements to decision-making processes, and robust theoretical foundation. Causal inference explanation, a model-agnostic post-hoc method, elucidates causal relationships between input features and model outputs, regardless of the model's architecture. It includes causal explanations, which clarify mechanisms and pathways~\cite{woodward2005making}, causal prediction, which forecasts future events based on established relationships~\cite{berk2008statistical} and causal intervention, which manipulates variables to observe effects~\cite{korb2004varieties,hagmayer2007causal}, supported by tools like structural causal models~\cite{pearl2009causality}.

Model-specific approaches predominantly include Tree Ensemble and Support Vector Machines in this era. Tree Ensemble approaches, provide interpretability through feature importance and decision paths~\cite{domingos2000mining}. In contrast, Support Vector Machines (SVMs) often require more intricate techniques, including rule extraction from support vectors and integration of training data with support vectors, to elucidate their decision-making processes~\cite{fu2004extracting}. They reflect the complexity and interpretability of different machine learning models.

Tree ensemble-based explanation methods can be categorized into path-based, feature-importance-based, and visualization-based approaches, each addressing distinct interpretability needs. Path-based explanations analyze classification decisions by tracing paths from root to leaf nodes. VFDT~\cite{domingos2000mining} enables real-time tree construction for high-speed data streams, and ROCR~\cite{sing2005rocr} provides visualizations like ROC curves for performance assessment. Enhancements by~\cite{ramos2005induction} optimize decision tree induction for complex datasets, improving the method's effectiveness. Feature-importance explanations evaluate feature significance through metrics such as split frequency or information gain. Random Forests~\cite{breiman2001random} aggregate predictions to enhance performance and reveal feature dependencies. DET (Density Estimation Tree)~\cite{ram2011density} extends decision trees for density estimation, offering improved interpretability. Visualization methods include decision path and feature importance visualizations. Tools like PaintingClass~\cite{teoh2003paintingclass} offer interactive tree construction and real-time adjustments~\cite{ankerst1999visual} to understand complex models.

Various post-hoc explainability techniques have been developed to enhance the interpretability of SVM models. Some methods involve integrating support vectors and hyperplanes, such as~\cite{fu2004extracting}, which constructs hyper-rectangles from their intersections. Another category combines training data with support vectors, including clustering approaches~\cite{nunez2002support}, hyper-rectangular rule extraction~\cite{zhang2005rule}, and multi-constraint optimization techniques~\cite{fung2005rule}. Lastly, multi-kernel SVM methods, like those proposed by~\cite{chen2007multiple} and~\cite{navia2006support}, employ multi-kernel approaches for feature selection and rule extraction or define linear rules using a growing SVC in Voronoi tessellation space. These methods differ in complexity and the degree of interpretability they provide.
\hypertarget{se4}{}
\section{The era of Deep Learning}
As shown in Table~\ref{tab3}, in this era, the shift from traditional machine learning—centered on manual feature engineering—to deep neural networks brings new challenges in interpretability. MXAI focuses on making the \enquote{black box} nature of deep models more transparent. Key efforts include balancing model performance with interpretability and developing both local and global explanation techniques to provide insights into model decisions and overall behavior, enhancing the trustworthiness of systems.
\begin{table*}[!t]
\centering
  \caption{Summary of MXAI methods in the deep learning era}
  \label{tab3}
\begin{tabular}{c|c|c|cc}
\hline
\multirow{14}{*}{\begin{tabular}[c|]{@{}l@{}}\makecell{The Era of Deep Learning \\ (2011-2016)}\end{tabular}} & \multirow{2}{*}{\hyperlink{Premodel}{Data explainability}} & Data quality analysis & \multicolumn{2}{c}{\cite{groves2011linked,wilderjans2011simultaneous,khaleghi2013multisensor,liberman2014new,RN142,tmazirte2013dynamical,acar2011scalable}} \\ \cline{3-5} 
 &  & Data interaction analysis & \multicolumn{2}{c}{\cite{vinyals2015show,chen2017multimodal,venugopalan2015sequence,sun2015human} } \\ \cline{2-5} 
 & \multirow{6}{*}{\hyperlink{In-model}{Model explainability}} & Inherently interpretable models & \multicolumn{2}{c}{\cite{ustun2016supersparse,lakkaraju2016interpretable,jung2017simple,alonso2015interpretability,lou2013accurate}} \\ \cline{3-5} 
 &  & \multirow{2}{*}{Deep neural network interpretability} & \multicolumn{1}{c|}{Decomposability} &{\cite{erhan2009visualizing,simonyan2013deep,nguyen2016synthesizing}}  \\ \cline{4-5} 
 &  &  & \multicolumn{1}{c|}{Algorithmic transparency} & {\cite{zeiler2011adaptive,zeiler2014visualizing,mahendran2015understanding,yosinski2015understanding,nguyen2016multifaceted,li2015convergent,koh2017understanding,shwartz2017opening}} \\ \cline{3-5} 
 &  & \multirow{3}{*}{Explain the training process} & \multicolumn{1}{c|}{Attention-based networks} &{\cite{vaswani2017attention,xiao2015application,lu2016hierarchical,das2017human}}  \\ \cline{4-5} 
 &  &  & \multicolumn{1}{c|}{Disentangled representations} & {\cite{kingma2013auto,higgins2017beta,chen2016infogan}} \\ \cline{4-5} 
 &  &  & \multicolumn{1}{c|}{Generate explanations} &\cite{antol2015vqa,hendricks2016generating,fukui2016multimodal,ross2017right}  \\ \cline{2-5} 
 & \multirow{6}{*}{\hyperlink{Postmodel}{Post-hoc explainability}} & \multirow{2}{*}{Multi-layer neural networks} & \multicolumn{1}{c|}{Model simplification} &\cite{che2016interpretable,thiagarajan2016treeview}  \\ \cline{4-5} 
 &  &  & \multicolumn{1}{c|}{Feature-related explanations} &\cite{montavon2017explaining,shrikumar2016not,sundararajan2017axiomatic,kindermans2017learning}  \\ \cline{3-5} 
 &  & \multirow{2}{*}{Convolutional Neural Networks} & \multicolumn{1}{c|}{\begin{tabular}[c]{@{}l@{}}\makecell{Understand decision-making processes}\end{tabular}} &\cite{zeiler2011adaptive,zeiler2014visualizing,bach2015pixel}  \\ \cline{4-5} 
 &  &  & \multicolumn{1}{c|}{Investigate module function} &\cite{mahendran2015understanding,ribeiro2016should,zhou2016learning,selvaraju2017grad}  \\ \cline{3-5} 
 &  & \multirow{2}{*}{Recurrent Neural Networks} & \multicolumn{1}{c|}{Feature-related explanations} &{\cite{arras2017explaining,karpathy2015visualizing}}  \\ \cline{4-5} 
 &  &  & \multicolumn{1}{c|}{Local explanations} &\cite{wisdom2016interpretable,krakovna2016increasing}  \\ \hline
\end{tabular}
\end{table*}

\hypertarget{Premodel}{}
\subsection{Data explainability}
\subsubsection{Data quality analysis}
Data quality analysis evaluates and interprets data from various sources to ensure integrity, consistency, and reliability, thereby improving the performance of deep learning models~\cite{groves2011linked,wilderjans2011simultaneous}. Traditional approaches, such as additive noise models with Bayesian or maximum likelihood estimation, often overlook data correlations, limiting the effectiveness of fusion algorithms~\cite{khaleghi2013multisensor}. To enhance data handling, methods like optimally weighted averages are employed for specific datasets, such as weather models~\cite{liberman2014new}. However, challenges like conflicts and inconsistencies in multimodal data persist~\cite{RN142}. To address these issues, information theory-based approaches dynamically reconfigure systems to manage sensor data inconsistencies~\cite{tmazirte2013dynamical}. Additionally, first-order optimization techniques in weighted least squares are used to capture underlying data structures and reconstruct missing values, thereby enhancing the robustness of data integration methods~\cite{acar2011scalable}. These combined strategies improve the overall reliability and performance of deep learning models by ensuring high-quality data.
\subsubsection{Data interaction analysis}
Data interaction analysis is dedicated to understanding and explaining the interactions between different types and sources of data. Such analyses not only help to improve the performance of models but also enhance their transparency and trust. To enhance the accuracy and coherence of generated descriptions, Vinyals et al.~\cite{vinyals2015show} model the correspondence between images and text using an encoder-decoder structure with an attention mechanism. Chen et al.~\cite{chen2017multimodal} synchronize text, audio, and video features by timestamp, utilizing the temporal relationships in multimodal data. Venugopalan et al.~\cite{venugopalan2015sequence} explain the model's decision-making by learning feature representations between video and text, employing attention mechanisms and feature fusion. Sun et al.~\cite{sun2015human} address sequence alignment by proposing an efficient strategy based on sampling multiple video clips. 
\hypertarget{In-model}{}
\subsection{Model explainability}
\subsubsection{Intrinsic interpretable models
}
These methods involve selecting from a predefined set of inherently interpretable (white box) techniques~\cite{RN7}. For example, Ustun et al.~\cite{ustun2016supersparse} introduce supersparse linear integer models for optimizing medical scoring systems, providing a clear, interpretable framework for healthcare professionals. Similarly, Lakkaraju et al.~\cite{lakkaraju2016interpretable} propose interpretable decision sets that combine description and prediction, enhancing model transparency. In contrast, Jung et al.~\cite{jung2017simple} develop simple rule-based approaches for complex decisions, ensuring understandability even in intricate scenarios. Furthermore, Alonso et al.~\cite{alonso2015interpretability} review current trends in the interpretability of fuzzy systems, highlighting their intuitive rule-based structure. Lastly, Lou et al.~\cite{lou2013accurate} present accurate intelligible models with pairwise interactions, achieving high performance while maintaining interpretability through visualized feature interactions. Together, these works underscore the importance of model interpretability across various applications.

\subsubsection{Deep neural network interpretability
}
Given the complexity of deep models, interpretation efforts focus primarily on decomposability and algorithmic transparency.

For decomposability, methods for deep model data processing include unit response visualization~\cite{erhan2009visualizing}, deconvolutional networks~\cite{simonyan2013deep}, and CNN-specific neuron preference~\cite{nguyen2016synthesizing}.

For algorithmic transparency, approaches involve guiding model structure revisions through interpretability~\cite{zeiler2011adaptive, zeiler2014visualizing}, abstraction analysis of deep image representations~\cite{mahendran2015understanding,nguyen2016multifaceted,yosinski2015understanding}, and convergence studies on feature learning in deep networks~\cite{li2015convergent}. Additionally, Koh et al.~\cite{koh2017understanding} assess how training data affects models by evaluating the influence of individual data points without retraining. Shwartz-Ziv et al.~\cite{shwartz2017opening} use an information-theoretic framework to analyze deep network behavior, offering insights into internal representation evolution and improving training efficiency during diffusion phases.
\subsubsection{Explain the training process}
Attention-based networks effectively direct information flow by weighting inputs or internal features, excelling in tasks like non-sequential natural language translation~\cite{vaswani2017attention}, fine-grained image classification~\cite{xiao2015application}, and visual question answering~\cite{lu2016hierarchical}. While attention units are not explicitly designed for human-readable explanations, they provide insights into information traversal within the network. And Das et al.~\cite{das2017human} adopt datasets simulating human attention to evaluate models by replicating human cognitive processes.

Disentangled representations improve interpretability by isolating individual causal factors within data. Variational autoencoders (VAEs) optimize models to match input distributions using information-theoretic metrics~\cite{kingma2013auto,higgins2017beta}, while InfoGAN~\cite{chen2016infogan} reduces feature entanglement through generative adversarial networks. Specialized loss functions in feedforward networks further facilitate feature disentanglement.

Additionally, deep networks can be trained to generate human-understandable explanations. Systems for tasks like visual question answering~\cite{antol2015vqa} and fine-grained image classification~\cite{hendricks2016generating} integrate explanation generation, providing multimodal outputs such as attention maps and textual interpretations~\cite{fukui2016multimodal}. Some methods adopt attention mechanisms to align network behavior with the desired explanations~\cite{ross2017right}.
\hypertarget{Postmodel}{}
\subsection{Post-hoc explainability}
\subsubsection{Multi-layer neural networks}
Multi-layer neural networks (MLPs), recognized for their capacity to model complex relationships, have driven advancements in interpretability techniques such as model simplification, feature relevance estimation, text and local interpretation, and model visualization~\cite{RN126}. For instance,~\cite{che2016interpretable} introduces interpretable mimic learning, a knowledge-distillation method that uses gradient boosting trees to create interpretable models with performance comparable to deep learning models. Similarly, Che et al.~\cite{thiagarajan2016treeview} present Treeview to enhance interpretability by partitioning the feature space with a tree structure. Feature relevance methods are employed to streamline complex models, thereby improving efficiency, robustness, and trust in predictions~\cite{montavon2017explaining,shrikumar2016not}. Additionally, theoretical validation efforts for multilayer neural networks have made progress in model simplification, nonlinear feature correlation, and hidden layer interpretation~\cite{sundararajan2017axiomatic,kindermans2017learning}.
\subsubsection{Convolutional Neural Networks}
Research on CNNs interpretability primarily focuses on two methods: mapping inputs to outputs to elucidate decision-making processes and examining how intermediate layers perceive the external world.

The first approach involves reconstructing feature maps from selected layers to understand activation effects. For instance, Zeiler et al.~\cite{zeiler2011adaptive} reconstruct high activations to reveal which parts of an image trigger specific responses, with further visualization provided by~\cite{zeiler2014visualizing}. Bach et al.~\cite{bach2015pixel} employ heatmaps and Layer-wise Relevance Propagation (LRP) to illustrate the contribution of each pixel to prediction outcomes, using Taylor series expansion around the prediction point for better accuracy.

Another approach focuses on intermediate layer interpretations. Mahendran et al.~\cite{mahendran2015understanding} develop a framework to reconstruct images within CNNs, showing that various layers maintain distinct levels of geometric and photometric consistency. Nguyen et al.~\cite{nguyen2016synthesizing} introduce a Deep Generator Network (DGN) to generate images that best represent specific output neurons in a CNN. LIME~\cite{ribeiro2016should} fits simple, interpretable models to approximate the behavior of target models near specific data points, providing local interpretations. CAM~\cite{zhou2016learning} and GradCAM~\cite{selvaraju2017grad} visualize activation positions for different classes to understand the model's decision-making process.
\subsubsection{Recurrent Neural Networks}
CNNs' limitations in handling long-term dependencies in sequential data have led to the development of RNNs, which utilize their own outputs as subsequent inputs, making them well-suited for sequential data processing. Research on RNN interpretability can be divided into feature-related and local explanations.

For feature-related explanations, Arras et al.~\cite{arras2017explaining} clarify the predictive processes of RNNs in specific tasks, highlighting the importance of various features. Building on this, Karpathy et al.~\cite{karpathy2015visualizing} offer methods to visualize and understand the roles of these features within the network, thus deepening our comprehension of RNN internal mechanisms.

In terms of local explanations, Wisdom et al.~\cite{wisdom2016interpretable} enhance RNN interpretability through sequential sparse recovery, which focuses on sparse activations over time to achieve local interpretability. Additionally, Krakovna et al.~\cite{krakovna2016increasing} improve RNN interpretability by integrating Hidden Markov Models (HMMs), providing a structured framework for sequential interpretation. Together, these approaches advance our understanding of RNNs by combining local interpretability techniques with structured sequential models.
\hypertarget{se5}{}
\section{The era of Discriminative Foundation Models}
This era emphasizes large-scale pre-trained models based on Transformer as concluded in Table~\ref{tab4}. These models are trained by unsupervised or self-supervised methods and leverage transfer learning to excel across various tasks with minimal task-specific data. They mark a shift from earlier advancements in neural network architectures and supervised learning that characterized the previous Deep Learning era (2010-2016).
\begin{table*}[!t]
\centering
  \caption{Summary of MXAI methods in the discriminative foundation models era}
  \label{tab4}
\begin{tabular}{c|c|c|cc}
\hline
\multirow{14}{*}{\makecell{The Era of Discriminative\\Foundation Models (2017-2021)}} & \multirow{2}{*}{\hyperlink{Data understanding}{Data explainability}} & Analyse multimodal datasets & \multicolumn{2}{c}{\cite{srinivasan2020interweaving,noroozi2019multimodal,kafle2017analysis,cortinas2023toward,zeng2019emoco,zeng2020emotioncues,wang2021dehumor,wang2018towards,escalante2020modeling,khan2017co,siegert2018using,kang2019visual}} \\ \cline{3-5} 
 &  & Structural relationship construction & \multicolumn{2}{c}{\cite{narasimhan2018out,zhang2023m3gat,zhuo2019explainable,sun2023modeling,lully2018enhancing}} \\ \cline{2-5} 
 & \multirow{7}{*}{\hyperlink{Model transparency}{Model explainability}} & \multirow{3}{*}{Behavioral explanation} & \multicolumn{1}{c|}{Architecture-independent} &\cite{cao2020behind,hendricks2021probing,chen2020uniter,frank2021vision}  \\ \cline{4-5} 
 &  &  & \multicolumn{1}{c|}{Transformer based} &\cite{chefer2021transformer,chen2022beyond,chefer2021generic,barkan2023learning}  \\ \cline{4-5} 
 &  &  & \multicolumn{1}{c|}{CLIP based} &\cite{li2023clip,lin2024tagclip,gandelsman2023interpreting,guo2023black}  \\ \cline{3-5} 
 &  & \multirow{4}{*}{Structural transparency} & \multicolumn{1}{c|}{GNN-based} &\cite{han2023unveiling,narasimhan2018out,zhang2023m3gat}  \\ \cline{4-5} 
 &  &  & \multicolumn{1}{c|}{Knowledge Graph-based} &\cite{zhang2022multimodal,wang2018dkn,ai2018learning,bellini2018knowledge}  \\ \cline{4-5} 
 &  &  & \multicolumn{1}{c|}{Causal-based} &\cite{CMCIR,agarwal2020towards,niu2021counterfactual}  \\ \cline{4-5} 
  &  &  & \multicolumn{1}{c|}{Others} &\cite{cadene2019rubi,cao2019interpretable,wang2023lcm}   \\ \cline{2-3} \cline{4-5} 
 & \multirow{5}{*}{\hyperlink{Post-hoc explainability}{Post-hoc explainability}}  &  Counterfactual based & \multicolumn{2}{c}{\cite{RN144,chen2020counterfactual,niu2021counterfactual,pan2019question,kanehira2019multimodal,hendricks2018generating,guo2020non,chen2019counterfactual}} \\ \cline{3-5} 
 &  & Bias mitigation & \multicolumn{2}{c}{\cite{pena2020bias,manjunatha2019explicit,hendricks2018women,ramakrishnan2018overcoming}} \\ \cline{3-5} 
 &  & \multirow{3}{*}{Multimodal learning process explanation} & \multicolumn{1}{c|}{Multimodal representation} &\cite{sun2023modeling,liang2023multimodal} \\ \cline{4-5} 
 &  &  & \multicolumn{1}{c|}{Multimodal reasoning} &\cite{niu2021counterfactual,zhang2022multimodal}  \\ \cline{4-5} 
 &  &  & \multicolumn{1}{c|}{Visualization} &\cite{liang2022multiviz,aflalo2022vl}  \\ \hline
\end{tabular}
\end{table*}
\hypertarget{Data understanding}{}
\subsection{Data explainability}
\subsubsection{Analyse multimodal datasets}
Srinivasan et al.~\cite{srinivasan2020interweaving} propose a multimodal interaction approach that integrates direct manipulation, natural language, and flexible unit visualizations to enhance visual data exploration. Noroozi et al.~\cite{noroozi2019multimodal} simplify complex data through effective visualization and analysis. To address limitations in VQA datasets, Kafle et al.~\cite{kafle2017analysis} introduce the TDIUC dataset to evaluate VQA algorithms, balancing question types and revealing strengths and weaknesses. In tackling high-dimensional and multimodal data complexity, annotations are focused on rather than raw input data~\cite{cortinas2023toward}. Tools for analyzing multimodal datasets and understanding annotations are proposed~\cite{zeng2019emoco,zeng2020emotioncues,wang2021dehumor}. Additionally, some methods~\cite{wang2018towards,escalante2020modeling} address dataset noise and bias issues. Unsupervised methods~\cite{khan2017co,siegert2018using,kang2019visual} are employed to improve prediction performance by identifying key features and simplifying semantic interpretation.
    
\subsubsection{Structural relationship construction}
Narasimhan et al.~\cite{narasimhan2018out} and Zhang et al.~\cite{zhang2023m3gat} leverage graph convolutional networks for factual visual question answering, focusing on inferring object relationships in images to answer questions accurately. Building on this, Zhuo et al.~\cite{zhuo2019explainable} extend the approach by constructing scene graphs for individual video frames and linking them into a spatio-temporal graph through object tracking. This method facilitates the interpretation of action sequences by aligning them with human logical reasoning. In parallel, Sun et al~\cite{sun2023modeling} propose a hypergraph-induced multimodal-multitask (HIMM) network to enhance semantic understanding, integrating unimodal, multimodal, and multitask hypergraph networks. This approach complements the graph-based methods by addressing complex semantic relationships. Additionally, Lully et al.~\cite{lully2018enhancing} explore the use of knowledge graphs to improve recommendation explainability through unstructured textual descriptions, addressing the challenge of making recommendations more interpretable and contextually relevant.
\hypertarget{Model transparency}{}
\subsection{Model explainability}
\subsubsection{Behavioral explanation}

In the realm of model behavior explanations, methods are divided into architecture-dependent and architecture-independent categories. Architecture-dependent methods analyze the internal mechanisms and structure of the model, while architecture-independent methods focus on the relationship between inputs and outputs. For example, DIME~\cite{lyu2022dime} enables precise analysis of multimodal models by breaking them down into unimodal contributions (UCs) and multimodal interactions (MIs), and is adaptable across various modalities and architectures. Grad-CAM++~\cite{chattopadhay2018grad} provides fine-grained visual explanations by highlighting relevant input regions, and LIFT-CAM~\cite{jung2021towards} combines layer-wise relevance propagation with feature map activations for improved interpretability across different models. Additionally, some unimodal approaches can be extended to multimodal scenarios, maintaining independence from specific architectures~\cite{mudrakarta2018did,hofmann2022towards,asokan2022interpretability,petsiuk2018rise}. For architecture-dependent approaches, we focus primarily on methods related to Transformer and CLIP models.

Transformer models have significantly advanced NLP by identifying patterns in large datasets, but their internal workings remain opaque. Various interpretation methods have been developed, each with distinct advantages and limitations. Probing tasks provide a straightforward means to evaluate specific knowledge within Transformers. For instance, Cao et al.~\cite{cao2020behind} use probing tasks to assess patterns learned during pre-training, while Hendricks et al.~\cite{hendricks2021probing} apply these tasks to image-language Transformers using detailed image-sentence pairs. Although effective, probing tasks often lack the granularity needed for deeper insights. Ablation studies offer a more focused analysis. For example, Chen et al.~\cite{chen2020uniter} employ them to determine optimal pre-training setups, enhancing Transformer performance. Frank et al.~\cite{frank2021vision} introduce cross-modal input ablation to evaluate how well Transformers integrate diverse data sources. However, these studies may not fully clarify individual predictions. Token attribution methods, such as~\cite{chen2022beyond}, help detail the contributions of input tokens to final predictions, aiding in understanding task-specific decisions. Attention weight analysis, as explored by~\cite{chefer2021generic}, reveals how Transformers interpret cross-modal data, highlighting the importance of specific input regions. However, these methods may not consistently explain the rationale behind individual predictions. LTX~\cite{barkan2023learning} generates interpretable maps of key input regions using an interpreter model to emphasize influential areas in Transformer predictions. This visual approach enhances the understanding of decision-making processes, though its applicability across various tasks and models requires further exploration.

In the context of multimodal models like CLIP, which integrates image and text data, methods often focus on analyzing the alignment between modalities. Techniques such as visualization and input perturbation help reveal how the model processes tasks. CLIP Surgery~\cite{li2023clip} addresses issues like erroneous visualizations and noisy activations using structural surgery and feature surgery, respectively. TagCLIP~\cite{lin2024tagclip}, on the other hand, employs a \enquote{local-to-global} framework to enhance interpretability and classification capabilities. Additionally, Gandelsman et al.~\cite{gandelsman2023interpreting} investigate spatial localization in CLIP, analyzing its ability to identify specific objects within images. This work improves CLIP's robustness by removing irrelevant features and enhancing zero-shot image segmentation capabilities. Lastly, Guo et al.~\cite{guo2023black} focus on adapting vision-language models with limited data by simultaneously enhancing textual prompts and image feature adaptation, improving performance across various tasks.

\subsubsection{Structural transparency}
Recent methods leveraging GNNs and knowledge graphs have made significant strides. Han et al.~\cite{han2023unveiling} adopt GNN-based techniques to capture hierarchical structures in various modalities, revealing complex correlations within images and heterogeneous interconnections among them. Knowledge graph-based approaches offer highly interpretable reasoning processes by integrating multimodal data, such as text and images, and turning analogical reasoning to link prediction tasks~\cite{zhang2022multimodal}. These techniques have been effectively applied in recommender systems. For instance, DKN~\cite{wang2018dkn} integrates knowledge graphs into news recommendations, while other works~\cite{ai2018learning} combine collaborative filtering with knowledge graphs for product recommendations. Additionally, Bellini et al.~\cite{bellini2018knowledge} merge the advantages of knowledge graphs and self-encoders to deliver accurate and interpretable recommendation results. These approaches enhance transparency by providing clear insights into the reasoning behind recommendations and predictions. Researchers have proposed a new causal-based framework for event-level Visual Question Answering (VQA) tasks. This approach uses causal intervention operations to reveal the intricate causal structures linking visual and linguistic modalities~\cite{CMCIR}.

On the other hand, analyzing outputs from problem-branching networks and the master model helps users understand multimodal information fusion and unimodal bias~\cite{cadene2019rubi}. Conversely, Cao et al.~\cite{cao2019interpretable} adopt a dependency tree to enhance interpretability in VQA. TextLighT and VTCAM are proposed to reduce costs and improve transparency~\cite{wang2023lcm}.

\hypertarget{Post-hoc explainability}{}
\subsection{Post-hoc explainability}
\subsubsection{Counterfactual reasoning}
Counterfactuals aim to infer the causes of predictions and their relationships under input distortions. To enhance VQA model interpretability, Agarwal et al.~\cite{agarwal2020towards} reveal and reduce spurious associations using invariant and covariant editing, providing causal explanations and enhancing model transparency and reliability. Chen et al.~\cite{chen2020counterfactual} assess and improve model robustness by generating and analyzing counterfactual samples. Niu et al.~\cite{niu2021counterfactual} adopt counterfactuals for causal inference to analyze and explain language bias. Pan et al.~\cite{pan2019question} provide counterfactual examples allow users to investigate and understand how the VQA model produces its results and the underlying causes. In visual captioning, counterfactual interpretations emphasize observations leading to certain outcomes~\cite{kanehira2019multimodal,hendricks2018generating}. Counterfactual reasoning is also applied in reinforcement learning settings for non-autoregressive image captioning~\cite{guo2020non} and scene-graph representation~\cite{chen2019counterfactual}.

\subsubsection{Bias mitigation}
Unbalanced datasets or inappropriate feature selection can compromise input data quality and affect model fairness. Pena et al.~\cite{pena2020bias} identify sources of bias in multimodal systems through post-hoc analyses and explainable tools, offering improvements to mitigate these biases. Some methods~\cite{manjunatha2019explicit,hendricks2018women,ramakrishnan2018overcoming} analyze model behavior and performance with multimodal data, identify biases and suggest improvements using visualization tools and other methods.

\subsubsection{Multimodal learning process explanation}
We classify MXAI applications in the multimodal learning process into three categories: explainable multimodal representation, multimodal reasoning, and visualization.

In the category of explainable multimodal representation, Sun et al.~\cite{sun2023modeling} propose hypergraph networks to manage higher-order relationships within and between modalities and tasks in unimodal, multimodal, and multitasking scenarios. Liang et al.~\cite{liang2023multimodal} introduce a method to quantify multimodal interactions, highlighting how intermodal interactions enhance signal fusion.

For multimodal reasoning, Niu et al.~\cite{niu2021counterfactual} present CF-VQA, the first framework to treat linguistic bias in VQA as a causal effect, offering a causality-based explanation for VQA debiasing methods. Zhang et al.~\cite{zhang2022multimodal} transform the multimodal analogy reasoning task into a link prediction task.

In the visualization category, Liang et al.~\cite{liang2022multiviz} employ MULTIVIZ, a four-step process to visualize and explain multimodal model behavior. VL-InterpreT~\cite{aflalo2022vl} offers interactive visualizations to interpret attention and hidden representations.
\hypertarget{se6}{}
\section{The era of Generative Large Language Models}
This era focuses on advancing generative tasks by the foundations established by Discriminative Models (2017-2021). Unlike their predecessors, these models, such as GPT-4~\cite{openai2023gpt4}, BLIP-2~\cite{li2023blip} and their successors, generate coherent, contextually relevant text, enhancing explainability by providing natural language explanations for outputs. This advancement bridges the gap between human understanding and machine decision-making, enabling more nuanced interactions and insights into model behavior. We have concluded those works in Table~\ref{tab5}.
\begin{table*}[!t]
\centering
 \caption{Summary of MXAI methods in the generative large language models era}
\label{tab5}
\begin{tabular}{c|c|c|cc}
\hline
\multirow{18}{*}{\begin{tabular}[c]{@{}c@{}}The Era of Generative Large\\ Language Models\\ (2022-2024)\end{tabular}} & \multirow{3}{*}{\hyperlink{MMLs Data understanding}{Data explainability}} & Explaining datasets & \multicolumn{2}{c}{\cite{dibia2023lida,narayan2022can,huang2023benchmarking,li2024table,zhang2023generative,singh2024rethinking,bordt2024data}} \\ \cline{3-5} 
 &  & Data selection & \multicolumn{2}{c}{\cite{zhu2024multimodal,bruce2024genie,nguyen2022quality,deng2024robust}} \\ \cline{3-5} 
 &  & Graph modeling & \multicolumn{2}{c}{\cite{wei2024rendering,xu2024llm,huang2024multimodal}} \\ \cline{2-5} 
 & \multirow{8}{*}{\hyperlink{Explaining MLLMs}{Model explainability}} & \multirow{2}{*}{Process explanation} & \multicolumn{1}{c|}{Explain multimodal-ICL} & \cite{sun2024generative,baldassini2024makes,laurenccon2024obelics,awadalla2023openflamingo,zhang2024narr,shukor2023beyond,luo2024does} \\ \cline{4-5} 
 &  &  & \multicolumn{1}{c|}{Explaining multimodal-CoT} &\cite{ge2023chain,zhang2023multimodal,rose2023visual,himakunthala2023let}  \\ \cline{3-5} 
 &  & \multirow{2}{*}{Explainable data augmentation} & \multicolumn{1}{c|}{Robustness enhancement} &\cite{xu2022contrastive,tang2023does,khan2023q}  \\ \cline{4-5} 
 &  &  & \multicolumn{1}{c|}{Small models training} &\cite{wang2024t,whitehouse2023llm,an2023generalized}  \\ \cline{3-5} 
 &  & \multirow{4}{*}{Inherent interpretability} & \multicolumn{1}{c|}{Image-Text understanding} &\cite{li2023blip,liu2024llavanext,li2024llava}  \\ \cline{4-5} 
 &  &  & \multicolumn{1}{c|}{Video-Text understanding} &\cite{2023videochat,ma2023dolphins}  \\ \cline{4-5} 
 &  &  & \multicolumn{1}{c|}{Audio-Text understanding} &\cite{tang2023salmonn,chu2023qwen}  \\ \cline{4-5} 
 &  &  & \multicolumn{1}{c|}{Multimodal-Text understanding} &\cite{alayrac2022flamingo,yang2023mm,chen2023x,panagopoulou2023x,moon2023anymal}  \\ \cline{2-5} 
 & \multirow{7}{*}{\hyperlink{MMLs Post-hoc}{Post-hoc explainability}} & \multirow{2}{*}{Probing-based explanation} & \multicolumn{1}{c|}{Classifier-based} & \cite{tao2024probing,verma2024mysterious,qi2023limitation,li2022emergent} \\ \cline{4-5} 
 &  &  & \multicolumn{1}{c|}{Parameter-free} &\cite{sreeram2024probing,lu2024text,an2024agla}  \\ \cline{3-5} 
 &  & \multirow{3}{*}{Reasoning-based explanation} & \multicolumn{1}{c|}{Modular design-based} &\cite{xu2024active,chen2024reckoning}  \\ \cline{4-5} 
 &  &  & \multicolumn{1}{c|}{External world knowledge-based} &\cite{sreeram2024probing,lu2024text,yoneda2023statler}  \\ \cline{4-5} 
 &  &  & \multicolumn{1}{c|}{Feedback-based} &\cite{huang2022inner,wang2024describe,an2024knowledge}  \\ \cline{3-5} 
 &  & \multirow{2}{*}{Example-based explanation} & \multicolumn{1}{c|}{Counterfactual examples} &\cite{kim2024if,sreeram2024probing,li2024eyes,wu2023reasoning,zhang2024if,lewis2024using}  \\ \cline{4-5} 
 &  &  & \multicolumn{1}{c|}{Adversarial examples} &\cite{wang2024stop,cui2024robustness}  \\ \hline
\end{tabular}
\end{table*}
\hypertarget{MMLs Data understanding}{}
\subsection{Data explainability}
\subsubsection{Explaining datasets}
    LLMs can effectively explain datasets through interactive visualization and data analysis.  LIDA~\cite{dibia2023lida} generates grammar-agnostic visualizations and infographics to understand the semantics of the data, enumerate the relevant visualization goals, and generate visualization specifications. Other methods~\cite{narayan2022can,huang2023benchmarking,li2024table,zhang2023generative} enhance the explainability of the datasets by analyzing the datasets.

    By combining multimodal information and powerful natural language processing capabilities, LLMs can provide comprehensive, in-depth, customized, and efficient explanations of data~\cite{singh2024rethinking}. Bordt et al.~\cite{bordt2024data} explore the capabilities of LLMs in understanding and interacting with glass-box models, identifying unexpected behaviors, and suggesting repairs or improvements. The focus is on leveraging multimodal data interpretability to enhance these processes.

\subsubsection{Data selection}
Data selection is crucial in this era. It improves model performance and accuracy, reduces bias, enhances generalization, saves training time and resources, and boosts interpretability, making decision-making more transparent and aiding in model improvement~\cite{albalak2024survey}. Multimodal C4~\cite{zhu2024multimodal} enhances dataset quality and diversity by integrating multiple sentence-image pairs and implementing rigorous image filtering, excluding small, improperly proportioned images and those containing faces. This approach underscores text-image correlations, bolstering the robustness and interpretability of multimodal model training. A new generative AI paradigm based on heuristic hybrid data filtering is proposed to enhance user immersion and increase interaction levels between video generation models and language tools (e.g., ChatGPT~\cite{openai2022chatgpt})~\cite{bruce2024genie}. It enables the generation of interactive environments from individual text or image prompts.

In addition to the above, some works aim to improve the robustness of the model to distributional variations and out-of-distribution data~\cite{nguyen2022quality,deng2024robust}.
\subsubsection{Graph modeling}
    While MLLMs can process and integrate data from different modalities, they typically capture relationships implicitly. In contrast, graph modeling explicitly represents data nodes (e.g., objects in images, concepts in text) and their relationships (e.g., semantic associations, spatial relationships). This explicit representation facilitates a more intuitive understanding of complex data relationships. Some methods~\cite{wei2024rendering,xu2024llm,huang2024multimodal} combine graph structures with LLMs to improve both performance in complex tasks and model interpretability through multimodal integration.
\hypertarget{Explaining MLLMs}{}
\subsection{Model explainability}

\subsubsection{Process explanation}
In this era, the process explanation of MXAI emphasizes multimodal in-context learning (ICL) and multimodal Chain of Thought (CoT). The prominence of ICL comes from its ability to avoid extensive updates to numerous model parameters by using human-comprehensible natural language instructions~\cite{wang2023label}. Emu2~\cite{sun2024generative} enhances task-independent ICL by extending multimodal model generation. Link context learning (LCL)~\cite{tai2024link} focuses on causal reasoning to improve learning in Multimodal Large Language Models (MLLMs). A comprehensive framework for multimodal ICL (M-ICL) is proposed by~\cite{baldassini2024makes} in models like DEFICS~\cite{laurenccon2024obelics} and OpenFlamingo~\cite{awadalla2023openflamingo}, encompassing a wide range of multimodal tasks. MM-Narrator~\cite{zhang2024narr} utilizes GPT-4~\cite{openai2023gpt4} and multimodal ICL for generating audio descriptions (AD). Further advancements in ICL and new multimodal ICL variants are explored by~\cite{shukor2023beyond}. MSIER~\cite{luo2024does} uses a neural network to select instances that enhance the efficiency of multimodal context learning.
        
   Multimodal CoT addresses the limitations of single-modality models in complex tasks where relying solely on text or images fails to capture comprehensive information. Text lacks visual cues, and images miss detailed descriptions, limiting the model's reasoning abilities~\cite{RN176}. Multimodal CoT integrates and reasons across multiple data types, such as text and images~\cite{ge2023chain,zhang2023multimodal,rose2023visual,himakunthala2023let}. For instance, image recognition can be broken down into a step-by-step cognitive process, constructing a chain of networks that generate visual biases, which are added to input word embeddings at each step~\cite{ge2023chain}. Zhang et al.~\cite{zhang2023multimodal} first generate rationales from visual and language inputs, then combine these with the original inputs for reasoning. Hybrid rationales~\cite{zhou2024image} use textual rationale to guide visual rationale, merging features to provide a coherent and transparent justification for answers.
\subsubsection{Inherent interpretability}
  
In this subsection, we explore the intrinsic interpretability of Multimodal Large Language Models (MLLMs), focusing on two main categories of tasks: multimodal understanding and multimodal generation~\cite{RN177}. 

Tasks for multimodal understanding include image-text, video-text, audio-text, and multimodal-text understanding. In image-text understanding, BLIP-2~\cite{li2023blip} enhances interpretability by aligning visual and textual data through a two-stage pre-training process, which improves the coherence and relevance of image descriptions. LLaVA~\cite{liu2024visual} generates instruction-following data by converting image-text pairs into formats compatible with GPT-4~\cite{openai2023gpt4}, linking CLIP's visual encoder with LLaMA's language decoder, and fine-tuning them on visual language instruction datasets. Variants such as LLaVAMoLE~\cite{chen2024llava}, LLaVA-NeXT~\cite{liu2024llavanext}, and LLaVA-Med~\cite{li2024llava} build on this foundation, with enhancements tailored to specific domains and tasks.

For video-text understanding, unlike images, videos incorporate temporal dimensions, requiring models to process static frames and understand dynamic relationships between them. This adds complexity to multimodal models but also offers richer semantic information and broader application scenarios. VideoChat~\cite{2023videochat} constructs a video-centric instruction dataset emphasizing spatio-temporal reasoning and causal relationships. This dataset enhances spatio-temporal reasoning, event localization, and causal inference, integrating video and text to improve model accuracy and robustness. Dolphins~\cite{ma2023dolphins} combines visual and language data to interpret the driving environment and interact naturally with the driver. It provides clear, contextually relevant instructions, generates explanations for its suggestions, and continuously learns from new experiences to adapt to changing driving conditions.

For audio-text understanding, audio data, with its time-series nature, necessitates models that can parse and understand temporal dynamics. This extends the capabilities of multimodal understanding. Salmonn~\cite{tang2023salmonn} integrates a pre-trained text-based LLM with speech and audio encoders into a unified multimodal framework. This setup allows LLMs to process and comprehend general audio inputs directly, enhancing multimodal interpretability and providing insights into the relationship between text and audio data. Despite its strengths, Salmonn faces limitations in achieving comprehensive audio understanding. In contrast, Qwen-audio~\cite{chu2023qwen} advances this field by developing large-scale audio-language models. By leveraging extensive audio and text datasets, Qwen-audio improves the model's ability to process and interpret diverse auditory inputs, thus pushing the boundaries of multimodal comprehension and delivering robust performance across various audio-related tasks.

In multimodal-text understanding, models integrate and correlate information from various modalities (e.g., image, video, audio) with text. This involves comprehending each modality's features and aligning them effectively. For example, Flamingo~\cite{alayrac2022flamingo} merges diverse inputs to understand cross-modal correlations, addressing aspects such as temporal ordering and semantic consistency. However, it struggles with dense visual data and maintaining accuracy across modalities. To overcome these challenges, MM-REACT~\cite{yang2023mm} introduces a novel prompt design that represents text descriptions, spatial coordinates, and filenames for visual signals like images and videos. This design facilitates seamless multimodal integration and improves collaboration between text and vision models. Despite the strong zero-shot performance, MM-REACT may face difficulties with new visual contexts without further training. Building on Flamingo and MM-REACT, models such as X-LLM~\cite{chen2023x}, X-InstructBLIP~\cite{panagopoulou2023x}, and AnyMAL~\cite{moon2023anymal} aim to advance multimodal understanding and response generation. By enhancing multimodal integration, these models strive to process and interpret complex real-world information with greater accuracy and versatility.
\subsubsection{Explainable data augmentation}
Limited data availability often constrains model performance, making explainable data augmentation a viable solution. This section explores using LLMs to generate explanatory synthetic data. These methods fall into two main categories: robustness enhancement and guidance for training smaller models.

To enhance robustness, data augmentation helps create models less sensitive to specific features. Counterfactual data has been used in augmentation to improve model robustness~\cite{wang2021robustness}. LLMs can generate examples representing abnormal or rare situations, aiding smaller models in better generalizing to unseen data~\cite{xu2022contrastive,tang2023does}. Counterfactual thinking, a cognitive process where humans consider alternative realities, is employed in the Self-taught Data Augmentation (SelTDA) framework to generate pseudo-labels from unlabelled images, improving performance and robustness in Visual Question Answering (VQA) tasks~\cite{khan2023q}.

For training small models, explainable data augmentation also provides crucial guidance. Collecting high-quality Chain of Thought (CoT) rationale is time-consuming and costly, but approaches like T-SciQ~\cite{wang2024t} generate high-quality CoT reasoning as teaching signals, training smaller models for CoT reasoning in multimodal tasks. Whitehouse et al.~\cite{whitehouse2023llm} utilize LLMs, including ChatGPT~\cite{openai2022chatgpt} and GPT-4~\cite{openai2023gpt4}, to augment datasets, demonstrating the effectiveness of fine-tuning smaller multilingual models like mBERT and XLM-R with the synthesized data.
\hypertarget{MMLs Post-hoc}{}
\subsection{Post-hoc explainability}
\subsubsection{Probing based}
Probing-based explanations assess the internal representational capabilities of MLLMs by probing or testing them. These methods can be categorized into classifier-dependent and parameter-free probing methods.

Classifier-dependent probing methods evaluate a model's handling of global and local semantic representations using classifiers. For instance, Tao et al.~\cite{tao2024probing} assess the model's capability with semantic representations through classifier analysis. Similarly, Verma et al.~\cite{verma2024mysterious} examine the roles of modules in modelling visual attributes via classifiers. Prompt probing, involving visual, textual, and external information prompts, is crucial for understanding MLLMs' limitations in visual reasoning~\cite{qi2023limitation}. This probing indicates that nonverbal prompts may lead to catastrophic forgetting in MLLMs~\cite{li2022emergent}, diminishing their reasoning abilities.

In contrast, parameter-free probing methods do not use additional classifiers. DriveSim~\cite{sreeram2024probing} simulates various driving scenarios to evaluate MLLMs' performance in complex, dynamic environments. Additionally, the model's question answering ability across different contexts is assessed by altering page contexts without additional training~\cite{lu2024text}.
\subsubsection{Reasoning based}
We identify three primary methodologies for reasoning about architectures: reasoning with separate components, reasoning with external world knowledge, and reasoning with feedback.

Reasoning with separate components is crucial for advancing complex AI and robotic systems. Zeng et al.~\cite{zeng2022socratic} introduce Socratic models, which use independent modules for zero-shot multimodal reasoning, enhancing transparency. LM-Nav~\cite{shah2023lm} integrates large pre-trained models of language, vision, and action through modular processing, achieving high performance and interpretability. Similarly, embodied reasoning is explored by enabling language models to interact with robots, using feedback mechanisms to enhance reasoning capabilities~\cite{dasgupta2023collaborating}. This modularity allows for effective collaboration between components like language, perception, and action planning, improving system adaptability and performance.

Reasoning with external world knowledge involves integrating new evidence with existing knowledge. Conan~\cite{xu2024active}, an interactive environment, evaluates active reasoning through multi-round abductive inference, similar to Minecraft. It challenges agents to identify the necessary information for accurate answers. To enhance reasoning accuracy, RECKONING~\cite{chen2024reckoning} updates model parameters with contextual knowledge before questioning. Building on this, Statler~\cite{yoneda2023statler} employs state-maintaining language models to improve interpretation in dynamic settings, allowing for better contextual understanding.

Reasoning with feedback emphasizes continuous model adaptation based on environmental input~\cite{wang2024describe}. Inner Monologue~\cite{huang2022inner} allows models to iteratively plan actions and update internal states, enhancing reasoning. The training-free DKA~\cite{an2024knowledge} framework utilizes feedback from LLMs to specify the required knowledge. These advancements highlight the importance of feedback and contextual adaptation in enhancing model reasoning and overall performance.
\subsubsection{Example-based explanations}
Counterfactual explanations demonstrate the necessary changes to examples to significantly alter their predictions, providing insights into the model's decision-making process and helping explain individual predictions. For instance, counterfactual mechanisms have been used to reduce hallucinations and improve model trustworthiness~\cite{kim2024if}. Additionally, counterfactual reasoning in driving scenarios has been explored to enhance decision-making and understanding~\cite{sreeram2024probing}. By designing multimodal counterfactual tasks, researchers evaluate language models' reasoning capabilities and limitations, conclude that models often struggle with complex reasoning tasks, particularly when integrating text and image information~\cite{li2024eyes,wu2023reasoning,zhang2024if,lewis2024using}. This underscores the need for better multimodal explainability.

Adversarial examples reveal models' vulnerabilities through minor perturbations, potentially leading to mispredictions. For instance, a novel stop-reasoning attack has been introduced to bypass Chain of Thought (CoT) robustness, and the study also examines how CoT reasoning shifts when multimodal large language models (MLLMs) encounter adversarial images, providing insights into their reasoning under such conditions~\cite{wang2024stop}. Additionally, the response of large multimodal models to crafted perturbations has been investigated, revealing that while these models are generally robust, they remain vulnerable to adversarial attacks~\cite{cui2024robustness}. This highlights the need for improved interpretative techniques.
\hypertarget{eval}{}
\section{Evaluation datasets and metrics}
After reviewing the data, model, and post-hoc explainability of MXAI in various eras, we have organized the datasets and evaluation metrics in recent years as shown in Table~\ref{tab6}, with descriptions of the types of modalities, types of explanations, and so on, that they involve. We first categorize the relevant datasets based on different tasks into six types. And evaluation metrics are primarily classified into text explanation metrics, visual explanation metrics, and multimodal explanation metrics. 

For text explanation metrics in multimodal settings, metrics like BLEU-4, METEOR, ROUGE, CIDEr, and SPICE are commonly used. BLEU-4 measures n-gram precision, while METEOR focuses on word alignment, and ROUGE emphasizes recall. CIDEr and SPICE, in contrast, prioritize consensus and semantic matching, respectively. Additionally, GPT-Guided metrics~\cite{tang2024hawk} are used to evaluate responses by LLMs to assess the quality of answers. Moreover, human evaluation metrics provide an additional layer of credibility by offering reliable assessments of explanation quality and relevance\cite{park2018multimodal, 8481251}. Additionally, the NIST (National Institute of Standards and Technology) metric\cite{yan2023personalized} refines n-gram precision by accounting for the information gain of rarer n-grams, making it more sensitive to meaningful word choices. The Distinct metric\cite{yan2023personalized} promotes diversity in generated text by measuring the ratio of unique n-grams, ensuring that the outputs are varied and less repetitive.

For visual explanation metrics, Intersection over Union (IoU) is a widely used metric for evaluating visual explainability, especially in image segmentation and object detection tasks~\cite{rao2021first}. It calculates the ratio of the intersection to the union of the predicted and ground-truth regions. A higher IoU score reflects better alignment between the model’s visual explanation and the actual target, making it an effective metric for assessing the precision of visual interpretations. It can combine with textual explanation metrics to eval multimodal explanation.

For multimodal explanation metrics, CLIP Score~\cite{yan2023personalized} is a metric used to evaluate the alignment between generated images and their corresponding text prompts or original images in text-to-image or image-to-image tasks. It measures how well the generated output matches the input in terms of semantic relevance. MMEval~\cite{du2024uncovering} aligns with human preferences for assessing Causation Understanding of Video Anomalies (CUVA) and facilitates the evaluation of LLMs' ability to comprehend the causes and effects of video anomalies.
\begin{table*}[!t]
\caption{Summary of evaluation datasets and metrics}
\label{tab6}
\centering
\resizebox{\linewidth}{!}{ 
\begin{tabular}{c|c|c|c|c|c}
\hline
\textbf{Task} & \textbf{Dataset} & \textbf{Modalities} & \textbf{Metrics} & \textbf{Tags} & \textbf{Year} \\ \hline
\multirow{6}{*}{VQA} & VQA-X~\cite{park2018multimodal} & Image, Text & BLEU-4, METEOR, ROUGE, CIDEr, SPICE, Human eval & Textual explanation & 2018 \\ \cline{2-6} 
 & TextVQA-X~\cite{rao2021first} & Image, Text & Intersection over Union (IoU), BLEU-4, METEOR, ROUGE, CIDEr & Textual and visual explanation & 2021 \\ \cline{2-6} 
 & DocVQA~\cite{mathew2021docvqa} & Image, Text & Average Normalized Levenshtein Similarity (ANLS), Accuracy & Textual explanation & 2021 \\ \cline{2-6} 
 & SCIENCEQA~\cite{lu2022learn} & Image, Text & BLEU-1/4, ROUGE-L, and (sentence) Similarity, Human eval & Textual explanation & 2022 \\ \cline{2-6} 
 & LoRA~\cite{NEURIPS2023_617ff527} & Image, Text & Accuracy & Textual explanation & 2023 \\ \cline{2-6} 
 & PCA-Eval~\cite{chen2023towards} & Image, Text & Perception Score, Cognition Score, Action Score & Textual explanation & 2024 \\ \hline
Activity Recognition & ACT-X~\cite{park2018multimodal} & Image, Text & BLEU-4, METEOR, ROUGE, CIDEr, SPICE, Human eval & Textual explanation & 2018 \\ \hline
\multirow{3}{*}{Video Caption} & BDD-X~\cite{kim2018textual} & Text, Video & BLEU-4, METEOR, CIDEr-D, Human eval & Textual explanation & 2018 \\ \cline{2-6} 
 & Rank2Tell~\cite{Sachdeva_2024_WACV} & Text, Video & BLEU-4, METEOR, ROGUE, CIDER & Textual explanation & 2024 \\ \cline{2-6} 
 & VAST~\cite{chen2024vast} & Text, Video, Aideo & CIDER & Textual explanation & 2024 \\ \hline
\multirow{2}{*}{Visual Commonsense Reasoning} & VCR~\cite{zellers2019recognition} & Image, Text & Accuracy & Textual explanation & 2019 \\ \cline{2-6} 
 & DD-VQA~\cite{zhang2024common} & Image, Text & BLUE, CIDEr, ROUGE, METEOR, SPICE & Textual explanation & 2024 \\ \hline
\multirow{3}{*}{Recommendation tasks} & ExpFashion~\cite{lin2019explainable} & Image, Text & ROUGE, BLEU & Textual explanation & 2021 \\ \cline{2-6} 
 & REASONER~\cite{chen2024reasoner} & Text, Video & ROUGE, BLEU & Textual explanation & 2023 \\ \cline{2-6} 
 & Yan et al.~\cite{yan2023personalized} & Image, Text & BLEU, METEOR, NIST, Dinstinct, CLIP-Score, BERT-Score & Textual and visual explanation & 2023 \\ \hline
\multirow{3}{*}{Video anomaly understanding} & CUVA~\cite{du2024uncovering} & Text, Video & MMEval & Textual and visual explanation & 2024 \\ \cline{2-6} 
 & UCA~\cite{yuan2024towards} & Text, Video & BLEU, METEOR, ROUGE-L, CIDER & Textual explanation & 2024 \\ \cline{2-6} 
 & HAWK~\cite{tang2024hawk} & Text, Video & BLEU, GPT-Guided eval (Reasonability, Detail, Consistency) & Textual explanation & 2024 \\ \hline
\end{tabular}
}
\end{table*}

\hypertarget{se8}{}
\section{Future challenges and directions}
\subsection{Eliminating the MLLMs hallucination with MXAI}
Current methods for mitigating hallucinations mainly involve recognizing patterns and penalizing overconfident tokens using counterfactual samples~\cite{kim2024if,zhang2024if,li2024eyes,RN173}. Although promising, these approaches have significant limitations, including challenges in automating sample generation, ensuring transparency in explanations, generalizing across different contexts, achieving computational efficiency, and incorporating user feedback. Addressing these issues is vital for further progress.

\subsection{Visual Challenges in MLLMs}
MLLMs like GPT-4o~\cite{openai2024} encounter significant challenges with high-dimensional visual data, including content understanding, detail recognition, and visual reasoning~\cite{tong2024eyes,wu2024v}. These models often miss critical visual details, leading to biased or incomplete outputs. Improving the processing of visual data, advancing cross-modal fusion techniques, and enhancing interpretability are crucial steps toward overcoming these challenges. Additionally, robust user feedback mechanisms can guide ongoing refinements, ensuring that MLLMs evolve to better meet user needs. Despite recent advancements in MLLMs, challenges like interpretability, computational power and inference speed across diverse contexts persist~\cite{kang2024crespr}. Addressing these issues will significantly improve the capabilities, reliability, and practicality of MLLMs, making them more effective across a wide range of applications.

\subsection{Alignment with human cognition}
Aligning MLMs with human cognition is essential for improving interpretability, transparency, and credibility. Since language primarily functions as a communication tool rather than a thought process, integrating multimodal information is crucial for enhancing model reasoning capabilities~\cite{fedorenko2024language}. Future research will focus on refining multimodal fusion, developing explanations that mirror human cognitive processes, and ensuring MLMs' effectiveness in real-world scenarios. Models that are more robust, comprehensible, and effective in addressing complex challenges will be proposed.
\subsection{MXAI without ground truths}
MLLMs encounter significant interpretability challenges, particularly with ground truths. The complexity of multimodal data complicates the establishment of ground truths due to varying expressions, structures, and semantics. Ground truths are often subjective and costly to define, with human annotation inconsistencies adding to the challenge. Additionally, the opaque nature of deep learning models further complicates aligning outputs with ground truths. For example, cross-modal reasoning, crucial for tasks like visual question answering, requires integrating information from multiple modalities, making validation of ground truths even more complex. Evaluating MLLMs in dynamic environments, such as autonomous driving, introduces further difficulties as models must accurately process and interpret rapidly changing information. Addressing these challenges demands standardized datasets and benchmarks, improved model transparency, advanced cross-modal fusion techniques, and focused research on MLLMs interpretability relative to ground truths.
\section{Conclusion}
This paper categorizes Multimodal Explainable AI (MXAI) methods across four historical eras: traditional machine learning, deep learning, discriminative foundation models, and generative large language models. We analyze MXAI’s evolution in terms of data, model, and post-hoc explainability, and review relevant evaluation metrics and datasets. Looking ahead, key challenges include scaling explainability techniques, balancing model accuracy with interpretability, and addressing ethical concerns. The ongoing advancement of MXAI is crucial for ensuring AI systems are transparent, fair, and trustworthy.
 \bibliographystyle{IEEEtran}
\bibliography{refs.bib}

\newpage

\end{document}